\documentclass[10pt,conference]{IEEEtran}
\IEEEoverridecommandlockouts
\usepackage{amsmath,amssymb,amsfonts}
\usepackage[ruled,linesnumbered]{algorithm2e}
\usepackage{graphicx}
\usepackage{textcomp}
\usepackage{float}
\usepackage{booktabs}
\usepackage{multirow}
\usepackage{array}
\usepackage{amssymb}
\usepackage{xcolor}
\def\BibTeX{{\rm B\kern-.05em{\sc i\kern-.025em b}\kern-.08em
    T\kern-.1667em\lower.7ex\hbox{E}\kern-.125emX}}

\begin{document}

\title{SWSC: Shared Weight for Similar Channel in LLM
\thanks{*Corresponding author.}
}
\author{\centering
\IEEEauthorblockN{Binrui Zeng, Yongtao Tang, Xiaodong Liu{*}, Xiaopeng Li}
\IEEEauthorblockA{
College of Computer Science and Technology,
National University of Defense Technology,
Changsha, China\\
\{zengbinrui, tangyt,  liuxiaodong, xiaopengli\}@nudt.edu.cn}
}

\maketitle

\begin{abstract}
Large language models (LLMs) have spurred development in multiple industries. However, the growing number of their parameters brings substantial storage and computing burdens, making it essential to explore model compression techniques for parameter reduction and easier deployment. We propose SWSC, an LLM compression method based on the concept of ``Shared Weight for Similar Channel''. It uses the K-Means clustering algorithm to cluster model weights channel-by-channel, generating clusters with highly similar vectors within each. A representative vector from each cluster is selected to approximately replace all vectors in the cluster, significantly reducing the number of model weight parameters. However, approximate restoration will inevitably cause damage to the performance of the model. To tackle this issue, we perform singular value decomposition on the weight error values before and after compression and retain the larger singular values and their corresponding singular vectors to compensate for the accuracy. The experimental results show that our method can effectively ensure the performance of the compressed LLM even under low-precision conditions.

\end{abstract}


\section{Introduction}
Large language models (LLMs), with their rich knowledge reserves and powerful language understanding and generation capabilities, have brought new development opportunities to numerous industries \cite{zhao2024revolutionizing,wang2024large,jiang2024surveylargelanguagemodels}. However, as the number of their parameters continues to increase, it leads to greater storage and computing pressure. Therefore, it is necessary to explore model compression techniques to reduce the number of LLM parameters, thus significantly reducing its storage requirements. The compressed LLM is more easily deployed on various devices and platforms, including resource-constrained devices. This enables LLMs to be more widely applied in practical scenarios, promoting the popularization and development of artificial intelligence technology. 

Currently, LLM compression technologies mainly cover four directions \cite{zhu2024survey}: quantization, pruning, knowledge distillation, and low-rank decomposition. Each direction focuses on a core issue: how to achieve a higher level of model compression on the premise of ensuring that the performance of LLMs is not significantly affected. However, low-bit model compression methods can have a substantial impact on the performance of LLMs. Therefore, if a new compression technology orthogonal to the existing compression technologies can be designed, it will surely inject new vitality into the field of LLM compression.

\begin{figure}
    \centering
    \includegraphics[width=1\linewidth]{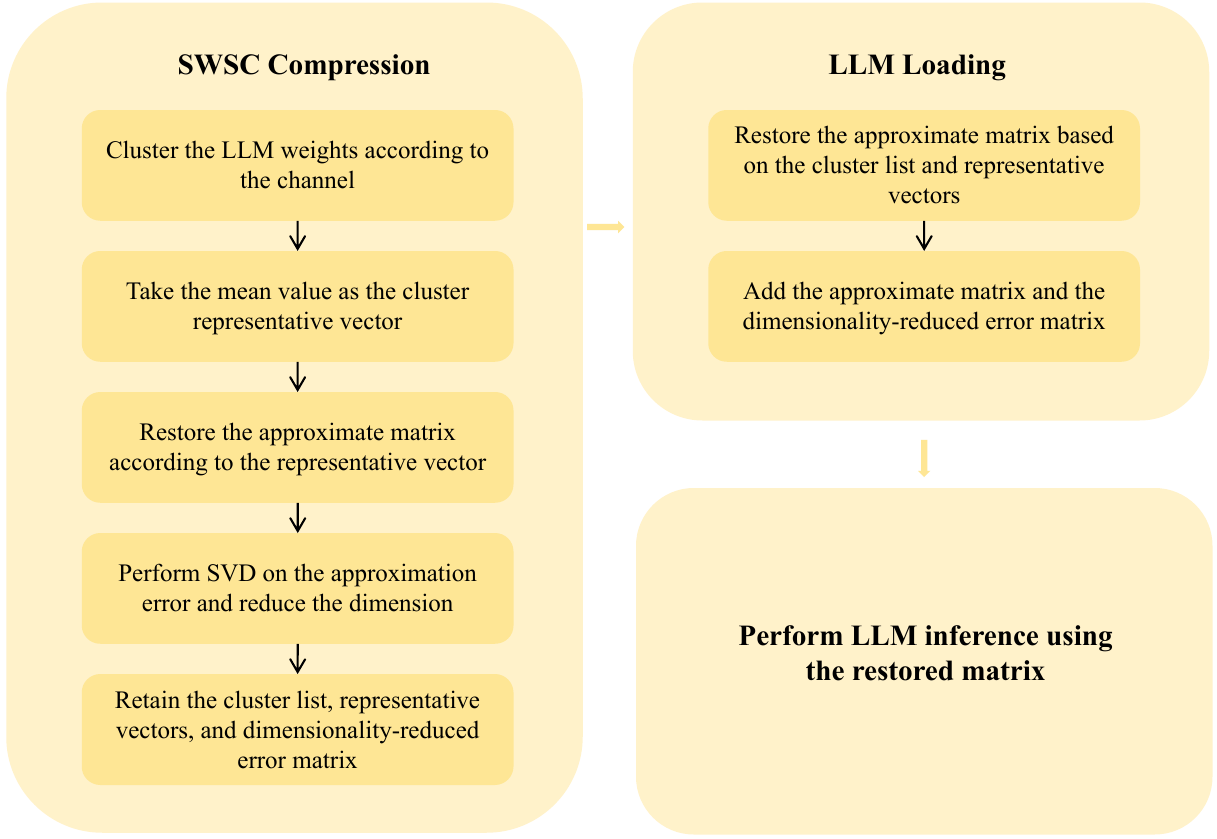}
    \caption{Flowchart of SWSC Compression and Restoration}
    \label{fig:swsc}
\end{figure}

To that end, we propose SWSC, an LLM compression method based on the concept of ``\textbf{S}hared \textbf{W}eight for \textbf{S}imilar \textbf{C}hannel''. Specifically, SWSC means adopting a set of shared parameters to replace the parameters corresponding to similar channels. Figure \ref{fig:swsc} shows the entire process of SWSC compression and restoration of LLM. This method uses the K-Means clustering algorithm to perform clustering operations on the model weights channel-by-channel, thus generating multiple clusters to ensure that the vectors within the same cluster have a high degree of similarity. Subsequently, a representative vector is selected from each cluster to approximately replace all the vectors in the cluster. In this way, when storing the weight parameters, only a cluster label list and the same number of representative vectors as the number of clusters need to be saved. This method significantly reduces the number of parameters of the model weights. In addition, given that outliers have a significant impact on the performance of LLM, to address the problem of outlier loss caused by the approximation step, we perform singular value decomposition \cite{von1947numerical} on the weight error values before and after compression. By retaining the larger singular values and their corresponding singular vectors, a dimensionality-reduction approximation of the error matrix is achieved. During the model inference stage, simply adding the approximated error matrix to the approximated model weights can obtain the weights used for inference. 

The main contributions of our work are as follows:
\begin{itemize}
    \item We propose a novel LLM compression technique based on clustering and singular value decomposition. This technique is orthogonal to existing model compression techniques and can significantly improve the weight compression efficiency of LLMs under the premise of ensuring performance.
    \item We solve the problem of outlier loss caused by compression by using singular value decomposition to correct the errors of the model before and after compression.
    \item The experimental results show that the method we proposed can effectively ensure that the performance of the compressed LLM is not significantly affected even under low-precision conditions.
\end{itemize}

\section{Related Work}
Current LLM compression techniques can be broadly categorized into four main types: quantization, pruning, Knowledge distillation, and low-rank decomposition. These classifications are grounded in distinct technical principles and operational approaches. In this section, we will present a concise overview and description of each of these four techniques. 

\subsection{LLM Quantization}
In the traditional representation of LLM weights, floating-point numbers are used. Quantization aims to compress parameters by converting them into integers or other discrete forms. Currently, the mainstream quantization techniques can be mainly divided into two types: Quantization Aware Training (QAT) and Post Training Quantization (PTQ). The main difference between these methods lies in when quantization is applied to compress the model.

QAT \cite{liu2023llmqatdatafreequantizationaware} enables LLMs to adapt to low precision representations during the training process, thus enhancing their ability to cope with the precision loss in the quantization process. In contrast, PTQ \cite{frantar2022gptq,xiao2023smoothquant,lin2024duquantdistributingoutliersdual} quantizes the parameters of LLMs after the training phase is completed.

\subsection{LLM Pruning}
In LLMs, some weight parameters are redundant and have a negligible impact on the model's performance. Pruning techniques, on the other hand, reduce the size or complexity of the model by removing these unnecessary or redundant components. According to different pruning units, pruning methods can be classified into two major categories: unstructured pruning and structured pruning. 

Unstructured pruning \cite{frantar2023sparsegpt, sun2024simpleeffectivepruningapproach} does not take into account the internal structure of LLMs. It simplifies LLMs by deleting specific parameters. The target of this method is individual weights or neurons in LLMs, usually by applying a threshold to zero out parameters below that threshold. This pruning method usually requires retraining LLMs, so it is relatively costly. Structured pruning \cite{ma2023llm, men2024shortgpt}, on the other hand, simplifies LLMs by removing entire structural units such as channels or layers. This method has the advantage of reducing model complexity and memory usage while keeping the overall structure of LLMs unchanged. 

\subsection{LLM Knowledge Distillation}
Knowledge Distillation (KD) transfers the knowledge from a complex model (the teacher model) to a simpler model (the student model), thereby improving the performance and generalization ability of the model while maintaining performance as much as possible. According to the access rights to the teacher model's predictions and parameters, KD can be divided into black-box KD and white-box KD. In the former, only the teacher's predictions are accessible, while in the latter, the teacher's parameters are available for use.

Black-box KD \cite{huang2022context,li2022explanationslargelanguagemodels,jiang2023lionadversarialdistillationproprietary} is applicable in situations where the teacher model is unknown or its internal parameters are inaccessible. For example, when using a commercial model or a pre-trained model provided by a third-party, only its input-output interface can be obtained. In such cases, black-box KD can be used to transfer knowledge to one's own student model. It is also suitable for scenarios such as model compression and deployment in mobile devices, embedded systems, etc., which have limited resources. On the other hand, white-box KD \cite{gu2024minillm,agarwal2024onpolicydistillationlanguagemodels} is suitable for the optimization and improvement of models. It is more appropriate when one has an in-depth understanding of the internal structure and parameters of the teacher model and hopes to further optimize the model through distillation. 

\subsection{LLM Low-rank Decomposition}
The core idea of low-rank decomposition is to decompose the high-dimensional parameter matrices in a LLM into the product form of low-rank matrices. In LLMs, the model parameters are usually represented in the form of matrices, and these matrices often have a very high dimension, resulting in high costs for model storage and computation. Through low-rank decomposition \cite{zhang2024loraprunestructuredpruningmeets,wu2023zeroquantfpleapforwardllms}, these high-dimensional matrices can be approximately represented as the product of two or more low-dimensional matrices, thus significantly reducing the number of model parameters while trying to maintain the model's performance. 

\section{Method}
In this section, we will elaborate on the specific operational procedures of using SWSC for LLM compression and compressed weight restoration.

\subsection{Motivation}
SWSC has drawn inspiration for LLM compression from the technique of Product Quantization \cite{5432202} (PQ). PQ is an efficient data compression and approximate nearest neighbor search technology. It divides the high-dimensional data space into multiple low-dimensional subspaces, performs clustering in each subspace respectively to form codebooks. 

When encoding data, the projections of the data in each subspace are matched with the corresponding codebooks, and the codeword indices are recorded. During decoding, the codewords are retrieved from the codebooks based on the indices and combined to restore the data. This technology is widely used in fields such as image retrieval and speech recognition, which can significantly reduce storage space and computational complexity while maintaining a good data approximation effect and improving data processing efficiency.

From the perspective of numerical analysis, the higher the degree of approximation between the approximate result and the original result, the smaller the corresponding loss. Through the implementation of channel-based clustering analysis on weights, it is found that under the condition of constant storage space, the mean square error of vectors in the same cluster is lower than that after RTN quantization, thereby demonstrating the feasibility of SWSC.
\begin{figure}[ht]
    \centering
    \includegraphics[width=1\linewidth]{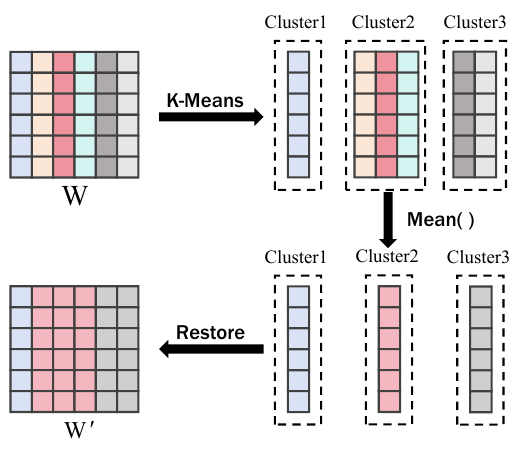}
    \caption{The process of clustering the weights of an LLM by channel and restoring them by taking the mean value.}
    \label{fig:kmeans}
\end{figure}

\subsection{Clustering-Driven Weight Transformation and Restoration}

SWSC first performs K-Means clustering on the weight matrix (assumed to be \(m\times m\)) to be compressed by channel. K-Means clustering is a classic unsupervised learning algorithm. Its basic idea is to divide data points into \(k\) clusters. Through continuous iteration to update the cluster centers and the belonging of data points, it aims to make the similarity of data points within the same cluster as high as possible and the similarity of data points between different clusters as low as possible. 

After the clustering operation is completed, vectors within the same cluster exhibit a high degree of similarity. In view of this, we use the mean value of the vectors within the cluster to approximately represent all the vectors in the entire cluster. In this way, when storing this weight, it is only necessary to store one cluster label vector with a dimension of \(1\times m\) and \(k\) (where \(k\) represents the number of clusters) representative vectors with a dimension of \(m\times1\). The elements in the cluster label vector correspond to the cluster numbers to which each channel belongs. By means of this method, a single weight of the LLM can be compressed from a size of \(m\times m\) to a size of \((k + 1)\times m\). Taking the self-attention layer of Llama-2-7B \cite{touvron2023llama} as an example, if the value of \(m\) for a certain weight is \(4096\) and \(k\) is set to \(256\), a compression rate of over \(90\%\) can be achieved on this matrix, significantly improving the storage efficiency. 

When it comes to restoring the model weights, one only needs to extract the corresponding vectors according to the pre-stored labels to complete the weight restoration task. Figure \ref{fig:kmeans} clearly and elaborately presents the entire process of weight transformation and restoration, which helps for an intuitive understanding of this crucial procedure.

\subsection{Error Compensation}

\begin{figure*}[ht]
    \centering
    \includegraphics[width=1\linewidth]{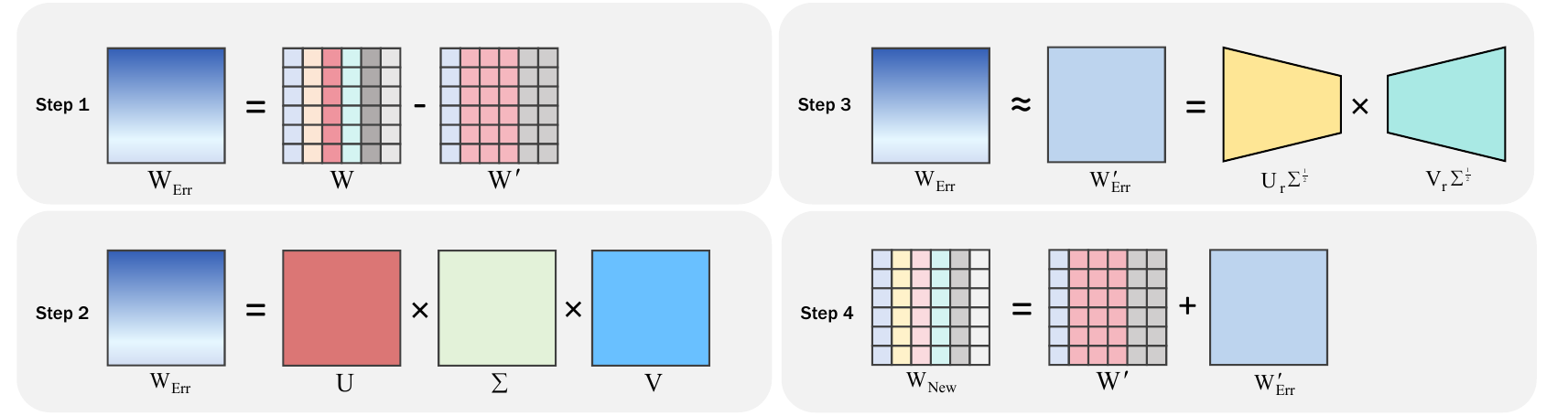}
    \caption{Compression Error Compensation Process. Among them, \(W\) and \(W'\) are the original matrix and the approximately restored matrix of the LLM respectively, and \(W_{Err}\) is the error matrix between the two matrices. \(U\), \(\Sigma\), and \(V\) are the results of singular value decomposition of \(W_{Err}\). \(W_{Err}'\) is the approximate error matrix, and \(W_{New}\) is the matrix that needs to be restored during the final inference.}
    \label{fig:svd}
\end{figure*}

After approximately restoring the weights, it will inevitably have a certain impact on the performance of the model. To mitigate this issue, we designed a compression error compensation method, as shown in Figure \ref{fig:svd}.

In the first step, after completing the approximate restoration of the matrix, we need to calculate the error matrix \(W_{Err}\) between the original matrix \(W\) and the restored matrix \(W'\). This error matrix will serve as the basis for subsequent error compensation operations, providing a crucial reference for further optimizing the model and improving its accuracy.

In the second step, we need to perform singular value decomposition on this error matrix. Singular value decomposition will generate three matrices: \(U\), \(\Sigma\), and \(V\). \(U\) is an orthogonal matrix, and its column vectors are called left singular vectors. Its column vectors form a set of orthonormal bases of the column space of the matrix. \(\Sigma\) is the singular value matrix, which is a diagonal matrix. The elements on its diagonal are called singular values, and the singular values are usually arranged in descending order. The number of non-zero singular values in this matrix is equal to the rank of the matrix. \(V\) is the right singular matrix, which is also an orthogonal matrix, and its column vectors are called right singular vectors. These column vectors form a set of orthonormal bases of the row space of the matrix. 

Subsequently, we achieve the dimensionality-reduction approximation of the error matrix by retaining the larger singular values and their corresponding singular vectors (assuming that the rank retained is r). The reason for taking this measure is mainly to reduce the storage space of the error-compensation matrix. Suppose that the matrices \(U\), \(\Sigma\), \(V\) are all \(m\times m\) matrices, the same as \(W\). After performing the dimensionality-reduction approximation operation, we no longer need to store these matrices with larger dimensions in their entirety. In fact, at this time, storing only one \(m\times r\) matrix \(U_{r}\Sigma^{\frac{1}{2}}\) and one \(r\times m\) matrix \(\Sigma^{\frac{1}{2}}V_{r}\) is sufficient to meet the subsequent compensation requirements. This change in the storage method has significantly reduced the amount of data stored and effectively improved the storage efficiency. 

In the final step, when loading the LLM weights, we only need to sum the matrix \(W'\) with \(W_{Err}'\) (the product of \(U_{r}\Sigma^{\frac{1}{2}}\) and \(\Sigma^{\frac{1}{2}}V_{r}\)) to restore the compressed weights. The restored weights will be applied to the inference calculations of the LLM, ensuring that the model can perform efficient and accurate inference processes based on accurate data.

\section{Experiments}

\subsection{Experimental Setup}

\subsubsection{LLMs}
To verify the feasibility of our method, we conducted experiments on Llama-2-7B \cite{touvron2023llama}. This LLM is an open-source LLM developed by Meta. It performs excellently in various scenarios such as text generation and question-answering, and thus has received significant attention in the field of natural language processing. Its moderate parameter scale of 7B enables model compression experiments to fully explore the technical key points with reasonable consumption. 

\subsubsection{Benchmarks}
We select WikiText-2 \cite{merity2016pointer} as the dataset to test the perplexity of the compressed model. WikiText-2 is a dataset composed of Wikipedia articles. It encompasses a wide variety of topics, has a moderate scale, and is widely used in the field of natural language processing. Since it is a standard benchmark dataset with broad comparability, using it to test the model's perplexity can accurately measure the model's language understanding and text-generation capabilities after compression. 

\subsubsection{Baseline}
Since SWSC is a brand new LLM compression method, we choose the RTN quantization method as the baseline for comparison with the SWSC compression method. This is because SWSC can compress the memory of large model weights to the same magnitude as that after RTN quantization by defining the number of clustering clusters and the rank of retained singular vectors, making this comparison both reasonable and effective.

\subsection{Main Results}
We performed SWSC compression and RTN quantization operations on the query projector, the key projector, and both in the self-attention layer of the Llama-2-7B model, respectively, compressing the weights to the same size. We only perform SWSC compression on the Query Projector and Key Projector because these two Projectors are not sensitive to the values themselves, so the original behavior of the model can still be largely preserved through approximate methods. However, the Value Projector stores the specific features of the model and has a higher requirement for accuracy, so it is not compressed. 

\setlength{\tabcolsep}{5.5mm}
\begin{table}[ht]
    \renewcommand{\arraystretch}{1.25}
\centering
\caption{The perplexity results of the Llama-2-7B model compressed by SWSC and quantized by RTN on the WikiText2 dataset.}
\begin{tabular}{cccc}
\toprule
\textbf{Projector}                & \textbf{Method} & \textbf{Avg. Bits}           & \textbf{Perplexity}        \\ \midrule
                                  & \textbf{RTN}    &                              & 20.550                     \\
                                  & \textbf{SWSC}   & \multirow{-2}{*}{\textbf{3}} & \textbf{6.547}             \\ \cmidrule(l){2-4} 
                                  & \textbf{RTN}    &                              & 4958.396                   \\
\multirow{-4}{*}{\textbf{Q}}      & \textbf{SWSC}   & \multirow{-2}{*}{\textbf{2}} & \textbf{7.297}             \\ \midrule
                                  & \textbf{RTN}    &                              & \textbf{6.073}                      \\
                                  & \textbf{SWSC}   & \multirow{-2}{*}{\textbf{3}} & 6.502                      \\ \cmidrule(l){2-4} 
                                  & \textbf{RTN}    &                              & {\color[HTML]{CB0000} nan} \\
\multirow{-4}{*}{\textbf{K}}      & \textbf{SWSC}   & \multirow{-2}{*}{\textbf{2}} & \textbf{7.148}             \\ \midrule
                                  & \textbf{RTN}    &                              & 40.206                     \\
                                  & \textbf{SWSC}   & \multirow{-2}{*}{\textbf{3}} & \textbf{8.102}             \\ \cmidrule(l){2-4} 
                                  & \textbf{RTN}    &                              & 10490                      \\
\multirow{-4}{*}{\textbf{Q \& K}} & \textbf{SWSC}   & \multirow{-2}{*}{\textbf{2}} & \textbf{10.886}            \\ \bottomrule
\end{tabular}
\label{table:ppl}
\end{table}

Table \ref{table:ppl} presents the experimental results of different projectors (Q, K, Q \& K) under RTN and SWSC methods with varying average bits and their corresponding perplexity values.

For projector Q, the perplexity of RTN surges significantly when the average bits drop from 3 to 2, while SWSC maintains relative stability. In the case of projector K, RTN yields an abnormal ``nan'' value at 2 average bits. When considering projector Q \& K, both methods exhibit higher perplexity compared to single-projector scenarios; however, the perplexity of SWSC after quantization remains acceptable. In particular, there are instances where the SWSC method with 2 average bits outperforms the RTN method with 3 average bits, suggesting that the SWSC is more robust under certain conditions. Overall, SWSC demonstrates relatively better performance across most cases. 

\subsection{Calculation of Average Bits}
\setlength{\tabcolsep}{6mm}
\begin{table}[ht]
    \renewcommand{\arraystretch}{1.25}
\centering
\caption{Average Bits Table Corresponding to the Number of Weight Clusters in the Self-Attention Layer of Llama-2-7B and the Rank of Singular Vector Retention}
\begin{tabular}{cc|cc}
\toprule
\textbf{Cluster} & \textbf{Avg Bits.} & \textbf{K}   & \textbf{Avg Bits.} \\ \midrule
\textbf{128}          & 0.5                & \textbf{64}  & 0.5                \\
\textbf{256}          & 1                  & \textbf{128} & 1                  \\
\textbf{512}          & 2                  & \textbf{256} & 2                  \\ \bottomrule
\end{tabular}
\label{table:avgbit}
\end{table}

Table \ref{table:avgbit} shows the corresponding changes in the average bits of the weights in the self-attention layer of the Llama-2-7B model due to the variations in the number of clusters and the rank of retention of singular vectors.

For the Llama-2-7B model, whenever the number of clustering groups increases by 128 or the rank of retained singular vectors increases by 64, the average number of bits increases by 0.5.

\section{Conclusion}
This paper introduces SWSC, an innovative LLM compression method based on the concept of "Shared Weight for Similar Channel". Using the K-Means clustering algorithm and singular value decomposition, SWSC effectively reduces the number of model weight parameters while maintaining the performance of the compressed LLM even under low-precision conditions. Experimental results on the Llama-2-7B model show that SWSC outperforms the RTN quantization method in many cases, especially when the average bits are low. This indicates that SWSC is a promising approach for LLM compression, which can potentially alleviate the storage and computing burdens associated with LLMs and facilitate their wider deployment across various devices. 

In future work, we will continue to optimize the SWSC method and combine it with other orthogonal model compression methods to test the compression performance of SWSC. 

\bibliographystyle{IEEEtran}
\bibliography{refs}
\end{document}